\documentclass[12pt]{article}

\usepackage{amsmath}
\usepackage{tikz-cd}
\usepackage{amssymb}
\usepackage{graphicx}
\usepackage[hyperfootnotes=false,hidelinks]{hyperref}
\usepackage{geometry}
\usepackage{mathtools}
\usepackage{titlesec}
\usepackage{titling}
\usepackage[backend=bibtex,style=numeric,sorting=none]{biblatex}
\usepackage{etoolbox}   % ← tiny helper to patch \maketitle
\usetikzlibrary{arrows.meta,positioning}
\titleformat{\section}      {\normalfont\Large\bfseries\centering}{\thesection}{1em}{}
\titleformat{\subsection}   {\normalfont\large\bfseries\centering}{\thesubsection}{1em}{}
\titleformat{\subsubsection}{\normalfont\normalsize\bfseries\centering}{\thesubsubsection}{1em}{}

\titleclass{\subsubsubsection}{straight}[\subsubsection]
\newcounter{subsubsubsection}[subsubsection]
\renewcommand\thesubsubsubsection{\thesubsubsection.\arabic{subsubsubsection}}
\titleformat{\subsubsubsection}
  {\normalfont\normalsize\bfseries\centering}{\thesubsubsubsection}{1em}{}
\titlespacing*{\subsubsubsection}{0pt}{1.5ex plus 0.5ex minus .2ex}{1em}

\pretitle{%
  \begin{center}
    \vspace*{-0.42cm}
    \rule{\linewidth}{3pt}\\[\smallskipamount]
    \bfseries\LARGE
    \vspace{0.22cm}
}
\posttitle{%
    \\[\smallskipamount]
    \rule{\linewidth}{2pt}
  \end{center}
  \vspace{0.12cm}
}

\preauthor{\begin{center}\large}
\postauthor{\end{center}}

\predate{\begin{center}\large}
\postdate{\end{center}}
\makeatletter
\g@addto@macro\maketitle{\thispagestyle{empty}}
\makeatother

\title{Tensor Field Models}

\author{%
  \begin{minipage}[t]{0.46\textwidth}\centering
    Alexander Strunk\thanks{Corresponding author: \href{mailto:astrunk.research@evercot.ai}{astrunk.research@evercot.ai}}\\
    Evercot AI\\
  \end{minipage}%
  \hspace{1em}% ← horizontal gap (≈ original spacing)
  \begin{minipage}[t]{0.46\textwidth}\centering
    Roland Assam\\
    Evercot AI\\
  \end{minipage}%
\vspace{0.42cm}
}

\date{19 June 2026}

\AtBeginBibliography{\sloppy}
\begin{document}

\hypersetup{pageanchor=false}
\maketitle  % <-- now produces the same title page, “automatically”

\vspace{-0.80cm} % keeps your abstract snug
\begin{abstract}
\noindent
This paper introduces Tensor Field Models (TFMs), realization-level
Mathematical Structures in which a learned Operator maps a product of admissible
component-section families to a prescribed family of time-dependent tangent
sections on a Generative State Manifold. Analytic and dynamical restrictions
are encoded through the choice of admissible families rather than imposed by
the root definition. Constructed, component-separable, and Tensor Bundle TFMs
provide structured refinements of this common object. In the conditional
realizations considered here, a structured condition
\(c=(c_1,\ldots,c_n)\) is mapped componentwise to a reusable collection
\(\mathbf H_c=(H_{c_1}^{(1)},\ldots,H_{c_n}^{(n)})\). In the architectures
evaluated here, the component representations remain distinct and are combined
only by the Field Operator to produce the generated Vector Field. All learned
models are trained using Flow Matching. Experiments show that TFMs can improve
performance and that amortized sampling enabled by reusable condition
representations can accelerate generation.
\end{abstract}

\newpage
\hypersetup{pageanchor=true}
\setcounter{page}{1}

% Main content
\section{Introduction}
A Tensor Field Model (TFM) is a realization containing its geometric domains,
admissible input- and output-section families, parameter space, and
parameterized Field Operator:
\begin{align*}
  \hat{T}
  =
  \left(
    \mathcal M,\mathcal I,
    \{(\Omega_i,E_i,\mathcal H_i)\}_{i=1}^n,
    \mathcal V,\Phi,\mathfrak F
  \right)
\end{align*}
Here, \(\mathcal M\) denotes the Generative State Manifold, \(\mathcal I\)
denotes the time interval, and each \(\pi_i:E_i\to\Omega_i\) is a smooth Vector
Bundle. Furthermore, \(\operatorname{Sec}(E_i)\) represents the space of
sections of \(E_i\), while
\(\mathcal H_i\subseteq\operatorname{Sec}(E_i)\) denotes a prescribed family of
admissible sections.

\begin{figure}[ht!]
  \centering
  \includegraphics[width=0.6\textwidth]{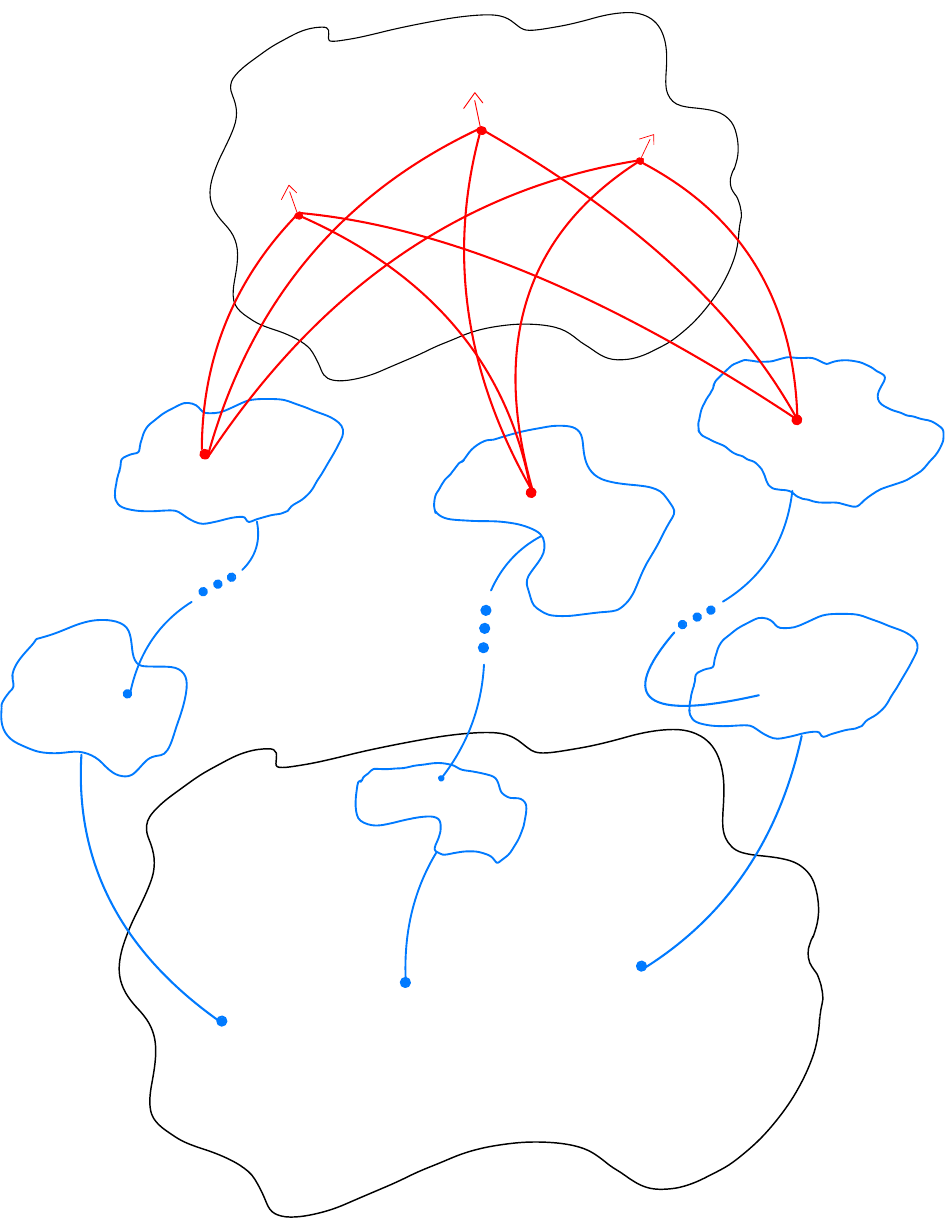}
  \caption{A Tensor Field Model can be visualized as a collection of blue
  component Fields on latent Bundles. Through the Operator
  \(\mathfrak F_\phi\), each admissible Field collection defines a red
  time-dependent tangent section on the Generative State Manifold
  \(\mathcal M\). This supports
  batched or parallel evaluation of independent queries while reusing the same
  Field collection.}
  \label{fig:Tensor_Field_Models}
\end{figure}

\noindent The prescribed output family satisfies
\(\mathcal V\subseteq
\operatorname{Sec}(\operatorname{pr}_{\mathcal M}^{*}T\mathcal M)\), and the
parameterized Field Operator has type
\begin{align*}
  \mathfrak F:
  \Phi\times\prod_{i=1}^n\mathcal H_i
  \longrightarrow
  \mathcal V
  \qquad
  \mathfrak F_\phi^{(n)}[\mathbf H]
  :=
  \mathfrak F(\phi,\mathbf H)
\end{align*}
Thus \(\mathfrak F_\phi^{(n)}[\mathbf H](x,t)\in T_x\mathcal M\).

\noindent Conditional realizations augment the Operator with a constructor
\(E_\theta:\mathcal C\to\prod_{i=1}^n\mathcal H_i\). When
\(\mathcal C=\prod_{i=1}^n\mathcal C_i\), this constructor may additionally
factorize componentwise.
\noindent For a constructed conditional TFM, the conceptual change is from directly learning
\(c\mapsto v_c\) in a high-dimensional section family to learning the
composition \(c\mapsto\mathbf H_c\mapsto\widehat v_c\). When the component Field
families have lower effective capacity than the original conditional
output family, their collection can provide a reduced representation.
This reduction is an architectural assumption, not a theorem following from
the existence of the factorization: without restrictions on the intermediate
families and Operator, an arbitrary conditional predictor can be written in
factorized form.
The model therefore does not treat each output-section evaluation as an
unrelated prediction. Instead, it reuses the same collection in all evaluations
under \(c\).

\noindent A condition-only construction is computationally useful because continuous-time
generative models repeatedly query a state- and time-dependent predictor under
fixed conditional information.
For \(N_q\) such queries, the component Fields are constructed once and the
point-evaluation Operator is evaluated \(N_q\) times. Whether this is faster
than an unfactorized predictor depends on the relative construction and
per-query costs analyzed in \autoref{sec:tfm-amortization}.

\begin{figure}[t]
\centering
\resizebox{0.98\textwidth}{!}{%
\begin{tikzpicture}[
  font=\small,
  node distance=6mm and 7mm,
  box/.style={draw, rounded corners, align=center, minimum height=9mm, inner sep=3.5pt},
  heavy/.style={box, line width=1.2pt, minimum width=27mm},
  query/.style={box, minimum width=43mm},
  arr/.style={-{Latex[length=2.0mm]}, thick}
]
  \node[box, minimum width=30mm] (condition) {structured condition\\$c=(c_i)_{i=1}^n$};
  \node[heavy, right=of condition] (constructor)
    {component constructors\\$E_{\theta_i}^{(i)}:\mathcal C_i\to\mathcal H_i$};
  \node[box, right=of constructor, minimum width=38mm] (fields)
    {Field collection\\$\mathbf H_c=E_\theta(c)=(H_{c_i}^{(i)})_{i=1}^n$};

  \node[query, right=10mm of fields, yshift=14mm] (f1)
    {query $q_1=(x_1,t_1)$\\$\widehat v_c(x_1,t_1)=F_\phi(\mathbf H_c,x_1,t_1)$};
  \node[query, right=10mm of fields] (f2)
    {query $q_2=(x_2,t_2)$\\$\widehat v_c(x_2,t_2)=F_\phi(\mathbf H_c,x_2,t_2)$};
  \node[query, right=10mm of fields, yshift=-14mm] (fk)
    {query $q_{N_q}=(x_{N_q},t_{N_q})$\\$\widehat v_c(x_{N_q},t_{N_q})=F_\phi(\mathbf H_c,x_{N_q},t_{N_q})$};

  \draw[arr] (condition) -- (constructor);
  \draw[arr] (constructor) -- (fields);
  \draw[arr] (fields.east) -- ++(4mm,0) |- (f1.west);
  \draw[arr] (fields.east) -- (f2.west);
  \draw[arr] (fields.east) -- ++(4mm,0) |- (fk.west);
  \node[above=1.5mm of f1] {point evaluations};
\end{tikzpicture}%
}
\caption{Point-evaluation formulation of a component-separable constructed
Tensor Field Model. The component constructors produce
\(\mathbf H_c=E_\theta(c)\). For \(N_q\) queries
\(q_j=(x_j,t_j)\) sharing the condition \(c\), the point-evaluation map
\(F_\phi\) reuses this collection to return
\(\widehat v_c(x_j,t_j)\in T_{x_j}\mathcal M\). The state \(x_j\) remains an
evaluation argument. In a dynamical specialization, each trajectory can
therefore be evaluated at its current state.}
\label{fig:tfm-overview}
\end{figure}
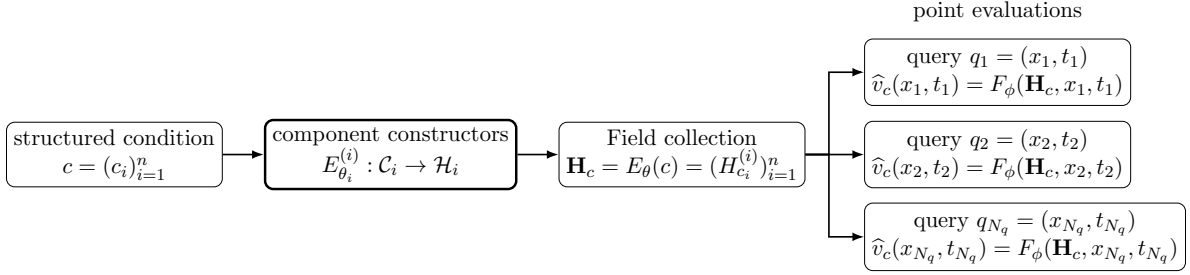

\newpage
\section{Mathematical Background}
\label{sec:tensor-field-background}

\subsection{Tensor Bundles and Tensor Fields}

A Tensor Bundle is a smooth Vector Bundle whose Fiber over each point is a
Tensor Product space formed from the corresponding Fiber of a specified Vector
Bundle and its Dual. More precisely, let \(M\) be a smooth \(d\)-dimensional
Manifold and let \(\pi_C:C\to M\) be a smooth real Vector Bundle of finite
rank \(q\). This carrier Bundle \(C\) supplies the Vector Spaces on
which the Tensors are formed and need not equal \(TM\). For
\(r,s\in\mathbb N_0\), the Tensor Bundle of type \((r,s)\) carried by \(C\)
is
\begin{gather*}
  \mathbb{T}^r_s(C)
  :=
  C^{\otimes r}\otimes(C^*)^{\otimes s}
  =
  \bigsqcup_{x\in M}\mathbb{T}^r_s(C)_x
  \\
  \mathbb{T}^r_s(C)_x
  :=
  C_x^{\otimes r}\otimes(C_x^*)^{\otimes s}
\end{gather*}
Here, \(C^*\to M\) is the Dual Bundle, and a zero-fold Tensor power is
understood to be the trivial line Bundle. The induced projection
\(\pi^r_s:\mathbb T^r_s(C)\to M\) sends every Tensor in
\(\mathbb T^r_s(C)_x\) to \(x\).

\begin{figure}[ht!]
  \centering
  \includegraphics[width=0.5\textwidth]{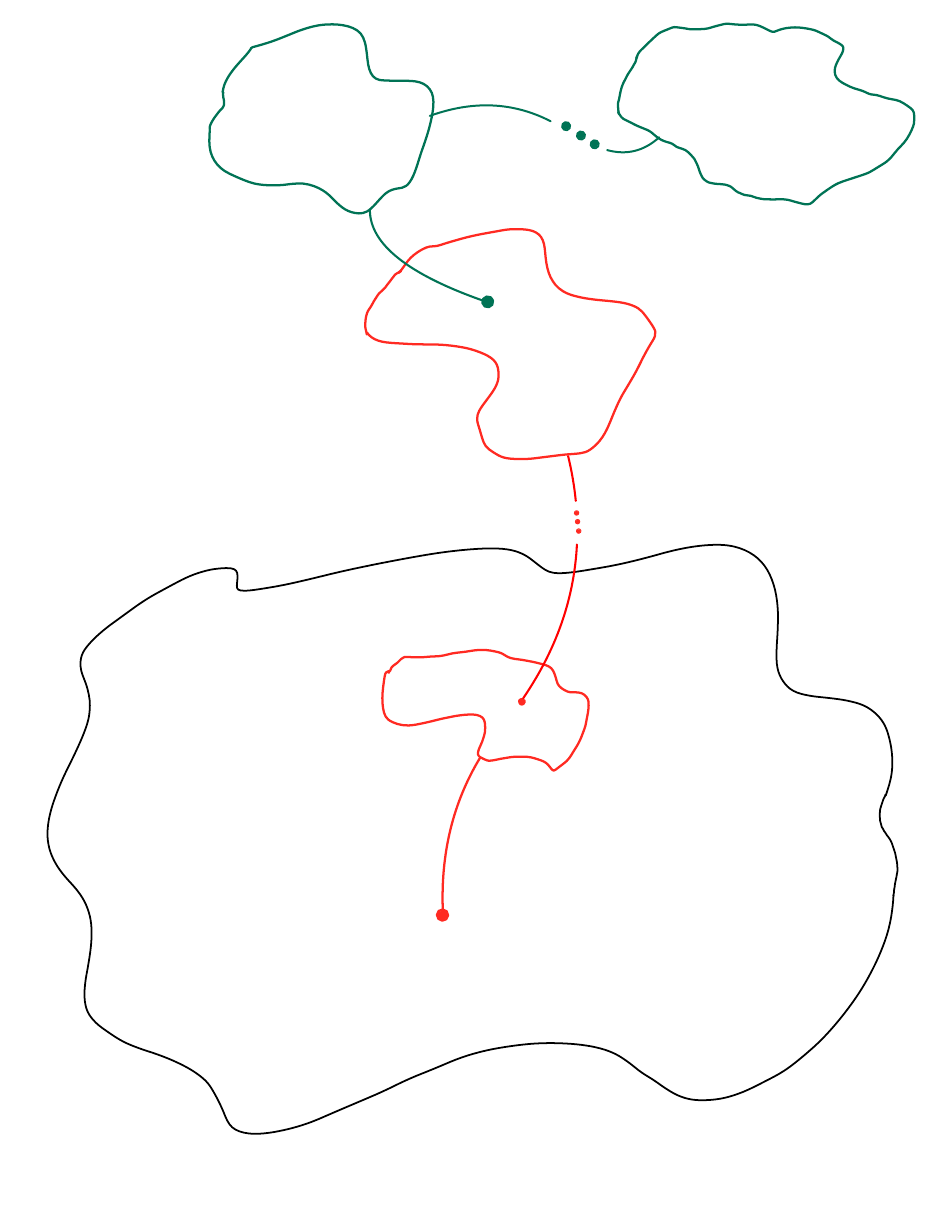}
  \caption{A Tensor Bundle induced by a carrier Bundle \(C\to M\). The lower
  region represents the Base Manifold \(M\), and the regions above selected
  base points represent Fibers \(\mathbb T^r_s(C)_x\). A smooth section
  assigns to every \(x\in M\) one Tensor \(A(x)\) in the corresponding
  Fiber. The carrier Fibers \(C_x\) need not be tangent spaces of \(M\).}
  \label{fig:Tensor_Field}
\end{figure}

\noindent A smooth \(C\)-Tensor Field of type \((r,s)\) is a smooth section of this Bundle:
\begin{align*}
  A\in\Gamma\!\left(\mathbb T^r_s(C)\right)
  \qquad
  \pi^r_s\circ A=\operatorname{id}_M
\end{align*}
Thus \(A\) denotes the Field, whereas
\(A(x)\in\mathbb T^r_s(C)_x\) denotes its value over \(x\). The word
``Tensor'' refers to the multilinear structure of the carrier Fiber \(C_x\).
It does not require \(C_x\) to be the tangent space \(T_xM\). The Base
Manifold \(M\) parametrizes the Fibers on which the Tensorial data live.

\noindent Let \(e_1,\ldots,e_q\) be a local frame of \(C\) over an open set
\(U\subseteq M\), and let \(e^1,\ldots,e^q\) be its dual frame. A
\(C\)-Tensor Field has the local representation
\begin{align*}
  A|_U
  =
  A^{a_1\ldots a_r}{}_{b_1\ldots b_s}(x)
  \,e_{a_1}\otimes\cdots\otimes e_{a_r}
  \otimes
  e^{b_1}\otimes\cdots\otimes e^{b_s}
\end{align*}
where the Fiber indices range from \(1\) to \(q\). If another frame is given
by \(\widetilde e_a=e_iG^i{}_a\), with
\(G:U\to\operatorname{GL}(q,\mathbb R)\), then the components satisfy
\begin{align*}
  \widetilde A^{a_1\ldots a_r}{}_{b_1\ldots b_s}
  ={}&
  (G^{-1})^{a_1}{}_{i_1}
  \cdots
  (G^{-1})^{a_r}{}_{i_r}
  G^{j_1}{}_{b_1}
  \cdots
  G^{j_s}{}_{b_s}
  A^{i_1\ldots i_r}{}_{j_1\ldots j_s}
\end{align*}
Repeated Fiber indices are summed. This frame-transformation law, rather than
the mere presence of several array indices, characterizes a geometric Tensor.
In general, \(G\) is a transition function of \(C\), not the Jacobian of a
coordinate change on \(M\). Moreover, the carrier rank \(q\) need not equal
the base dimension \(d\).

\subsubsection{Ordinary Tensors on the Base Manifold}

The familiar Tensor Bundles on \(M\) are recovered by choosing the carrier
Bundle \(C=TM\):
\begin{align*}
  \mathbb T^r_s M
  &:=
  \mathbb T^r_s(TM)
  &
  (\mathbb T^r_s M)_x
  &=
  (T_xM)^{\otimes r}\otimes(T_x^*M)^{\otimes s}
\end{align*}
For a coordinate chart \((U,x^1,\ldots,x^d)\), the coordinate frame
\(\partial_i=\partial/\partial x^i\) and its dual frame \(dx^j\) give
\begin{align*}
  T|_U
  =
  T^{i_1\ldots i_r}{}_{j_1\ldots j_s}(x)
  \,\partial_{i_1}\otimes\cdots\otimes\partial_{i_r}
  \otimes
  dx^{j_1}\otimes\cdots\otimes dx^{j_s}
\end{align*}
Under a coordinate change \(x\mapsto\widetilde x\), the general
frame-transformation law specializes to
\begin{align*}
  \widetilde T^{a_1\ldots a_r}{}_{b_1\ldots b_s}
  ={}&
  \frac{\partial\widetilde x^{a_1}}{\partial x^{i_1}}
  \cdots
  \frac{\partial\widetilde x^{a_r}}{\partial x^{i_r}}
  \frac{\partial x^{j_1}}{\partial\widetilde x^{b_1}}
  \cdots
  \frac{\partial x^{j_s}}{\partial\widetilde x^{b_s}}
  T^{i_1\ldots i_r}{}_{j_1\ldots j_s}
\end{align*}
Hence Scalar Fields \(f\in C^\infty(M)\), Vector Fields
\(X\in\Gamma(TM)\), One-Forms \(\omega\in\Gamma(T^*M)\), and Riemannian
metrics \(g\in\Gamma(\mathbb T^0_2 M)\) are the respective Base Manifold
examples of types \((0,0)\), \((1,0)\), \((0,1)\), and \((0,2)\).

\noindent For contrast, if \(C=M\times V\) is a trivial carrier Bundle, then a
\(C\)-Tensor Field is locally a smooth map from \(M\) into
\(V^{\otimes r}\otimes(V^*)^{\otimes s}\). Its indices are internal
\(V\)-indices and are unaffected by a change of coordinates on \(M\), although
they do transform under a change of frame in \(V\). Nontrivial carrier Bundles
give the corresponding construction without requiring a global frame. 

\newpage
\subsubsection{Mathematical Operations on Tensors}
Tensor Operations are defined fiberwise and, when the resulting pointwise
assignment is smooth, map Tensor Fields to Tensor Fields. Let \(C\to M\) be a
fixed carrier Bundle. The fundamental algebraic and functorial operations are
the following:
\begin{itemize}
  \item \textbf{Tensor Product}: For
  \(A\in\Gamma(\mathbb T^r_s(C))\) and
  \(B\in\Gamma(\mathbb T^u_v(C))\), their Tensor Product is the
  type-\((r+u,s+v)\) \(C\)-Tensor Field defined by
  \begin{align*}
    A\otimes B&\in\Gamma\!\left(\mathbb T^{r+u}_{s+v}(C)\right)
    & (A\otimes B)(x)&=A(x)\otimes B(x)
  \end{align*}
  Thus the operation concatenates the contravariant and covariant carrier
  slots of the two Tensors.

  \item \textbf{Contraction}: Contraction pairs one contravariant slot with
  one covariant slot by the natural Duality between \(C_x\) and \(C_x^*\).
  For \(r,s\geq1\), \(1\leq a\leq r\), and
  \(1\leq b\leq s\), contracting any selected pair gives
  \begin{align*}
    \operatorname{Ctr}_{a,b}:
    \Gamma\!\left(\mathbb T^r_s(C)\right)
    \longrightarrow
    \Gamma\!\left(\mathbb T^{r-1}_{s-1}(C)\right)
  \end{align*}
  For example, contraction of the first upper and lower indices has the local
  expression
  \begin{align*}
    (\operatorname{Ctr}_{1,1}T)^{i_2\ldots i_r}{}_{j_2\ldots j_s}
    =T^{k i_2\ldots i_r}{}_{k j_2\ldots j_s}
  \end{align*}
  The repeated Fiber index \(k\) is summed, and the result is independent of
  the chosen local frame.

  \item \textbf{Permutation}: The contravariant slots and the covariant
  slots may each be reordered without changing the Tensor type. For
  \(\sigma\in S_r\) and \(\tau\in S_s\),
  \begin{align*}
    (P_{\sigma,\tau}T)^{i_1\ldots i_r}{}_{j_1\ldots j_s}
    =T^{i_{\sigma(1)}\ldots i_{\sigma(r)}}{}
       _{j_{\tau(1)}\ldots j_{\tau(s)}}
  \end{align*}
  Linear combinations of such permutations yield, for example,
  symmetrization and antisymmetrization. For a covariant two-Tensor \(A\),
  \begin{align*}
    \operatorname{Sym}(A)_{ij}&=\tfrac12(A_{ij}+A_{ji})
    &
    \operatorname{Alt}(A)_{ij}&=\tfrac12(A_{ij}-A_{ji})
  \end{align*}
  An upper slot cannot be converted into a lower slot canonically because
  this would require a chosen identification \(C\cong C^*\), such as one
  induced by a Bundle metric.

  \item \textbf{Pullback of the Carrier Bundle}: Let \(f:L\to M\) be smooth.
  The pullback carrier Bundle \(f^*C\to L\) has Fiber
  \((f^*C)_x=C_{f(x)}\), and there is a canonical Bundle isomorphism
  \begin{align*}
    f^*\mathbb T^r_s(C)
    \cong
    \mathbb T^r_s(f^*C)
  \end{align*}
  Consequently, every \(A\in\Gamma(\mathbb T^r_s(C))\) induces
  \(f^*A\in\Gamma(\mathbb T^r_s(f^*C))\) with
  \((f^*A)(x)=A(f(x))\). This pullback is available for every Tensor type
  because it retains the pulled-back carrier \(f^*C\). It does not by itself
  identify that carrier with \(TL\).

  \item \textbf{Transport by a Bundle Map}: Let \(C\to L\) and \(D\to M\)
  be carrier Bundles, and let \(\Phi:C\to D\) be a fiberwise-linear Bundle
  map covering \(f:L\to M\). A covariant \(D\)-Tensor Field
  \(S\in\Gamma(\mathbb T^0_s(D))\) pulls back to a \(C\)-Tensor Field by
  \begin{align*}
    (\Phi^*S)_x(v_1,\ldots,v_s)
    =
    S_{f(x)}\bigl(\Phi_xv_1,\ldots,\Phi_xv_s\bigr)
    \qquad v_1,\ldots,v_s\in C_x
  \end{align*}
  A contravariant Tensor \(A_x\in C_x^{\otimes r}\) is pushed forward
  fiberwise:
  \begin{align*}
    \Phi_{*x}A_x
    =
    (\Phi_x)^{\otimes r}A_x
    \in D_{f(x)}^{\otimes r}
  \end{align*}
  In general this produces a Tensor Field along \(f\). If \(f\) is a
  diffeomorphism and \(\Phi\) is a Vector Bundle isomorphism, covariant and
  contravariant slots combine to transport mixed Tensor Fields.

  For ordinary Tensors on the Base Manifolds, the choice
  \(\Phi=df:TL\to TM\) recovers the usual pullback of covariant Tensors and
  pushforward of contravariant Tensors. For example,
  \begin{align*}
    (f^*\omega)_x(v)
    &=
    \omega_{f(x)}(df_xv),
    &
    f_{*x}X_x
    &=
    df_x(X_x).
  \end{align*}
\end{itemize}

For detailed treatments of these constructions, see
\cite{DifferentialGeometry,DifferentialGeometryManifoldConnections,
DifferentialGeometryManifold,SmoothManifolds,Manifolds,
GeometryTopologyPhysics}.
\newpage
\section{Tensor Field Models}
\label{sec:TFM}
A Tensor Field Model (TFM) with $n$ components is the following datum:
\begin{align*}
  \hat{T}
  =
  \left(
    \mathcal M,\mathcal I,
    \{(\Omega_i,E_i,\mathcal H_i)\}_{i=1}^n,
    \mathcal V,\Phi,\mathfrak F
  \right)
\end{align*}
For any Bundle \(\pi_E:E\to B\), the set of sections
\(\operatorname{Sec}(E)\) is defined as
\begin{align*}
  \operatorname{Sec}(E)
  &:=
  \left\{
    s:B\to E:
    \pi_E\circ s=\operatorname{id}_B
  \right\}
\end{align*}

\noindent The entries of \(\hat{T}\) satisfy the following typing conditions. The generative state space
\(\mathcal M\) is a smooth Manifold, and the time domain
\(\mathcal I\subseteq\mathbb R\) is a nondegenerate interval, viewed as a
smooth one-dimensional Manifold, possibly with boundary. For each
\(i=1,\ldots,n\), \(\Omega_i\) is a smooth Base
Manifold, \(\pi_i:E_i\to\Omega_i\) is a finite-rank smooth Vector Bundle, and
\(\mathcal H_i\) is a specified nonempty family of admissible component
sections:
\begin{align*}
  \varnothing
  \ne
  \mathcal H_i
  \subseteq
  \operatorname{Sec}(E_i)
\end{align*}
Define the product of admissible component sections by
\begin{align*}
  \mathfrak H^{(n)}
  :=
  \prod_{i=1}^n\mathcal H_i,
  \qquad
  \mathbf H=(H^{(1)},\ldots,H^{(n)})
  \in\mathfrak H^{(n)}
\end{align*}

\noindent Let \(\operatorname{pr}_{\mathcal M}:\mathcal M\times\mathcal I\to\mathcal M\)
denote projection onto the state, and let
\(\pi_{\mathcal M}:T\mathcal M\to\mathcal M\) be the Tangent Bundle
projection. The pullback Tangent Bundle has Fiber
\begin{align*}
  \left(\operatorname{pr}_{\mathcal M}^{*}T\mathcal M\right)_{(x,t)}
  \cong T_x\mathcal M
\end{align*}
The output datum is a specified nonempty admissible family
\begin{align*}
  \varnothing
  \ne
  \mathcal V
  \subseteq
  \operatorname{Sec}\!\left(
    \operatorname{pr}_{\mathcal M}^{*}T\mathcal M
  \right)
\end{align*}
There is a canonical identification
\begin{align*}
  \operatorname{Sec}\!\left(
    \operatorname{pr}_{\mathcal M}^{*}T\mathcal M
  \right)
  &\cong
  \left\{
    v:\mathcal M\times\mathcal I\to T\mathcal M:
    \pi_{\mathcal M}\circ v=\operatorname{pr}_{\mathcal M}
  \right\}
\end{align*}
Under this identification, an output is a time-dependent tangent Field \(v\)
satisfying \(v(x,t)\in T_x\mathcal M\).

\noindent \(\Phi\) is a specified nonempty parameter set, equipped with any topological
or smooth structure required by the application. The Field Operator datum
\begin{align*}
  \mathfrak F:
  \Phi\times\mathfrak H^{(n)}
  \longrightarrow
  \mathcal V
\end{align*}
is a parameterized Field Operator. No linearity, locality, or differentiability
is assumed unless it is separately imposed. For fixed
\(\phi\in\Phi\) and \(\mathbf H\in\mathfrak H^{(n)}\), write
\begin{align*}
  \mathfrak F_\phi^{(n)}[\mathbf H]
  :=
  \mathfrak F(\phi,\mathbf H)
\end{align*}
Pointwise evaluation is
\begin{align*}
  F_\phi(\mathbf H,x,t)
  &:=
  \bigl(\mathfrak F_\phi^{(n)}[\mathbf H]\bigr)(x,t)
  &
  F_\phi(\mathbf H,x,t)&\in T_x\mathcal M
\end{align*}

\subsection{Hierarchy of Tensor Field Models}
\label{sec:tfm-hierarchy}

The realization \(\hat{T}\) is the common root of the hierarchy. C-TFMs and
CS-TFMs form the construction branch by adding a constructor and then a
componentwise factorization of that constructor. TB-TFMs add Tensor Bundle
typing to the component Bundles. This typing can be combined with
either construction class when the corresponding Tensor specifications and
Bundle isomorphisms are supplied.

\subsubsection{Construction Hierarchy}

Let \(\mathcal C\) be a specified nonempty condition set and \(\Theta\) a
specified nonempty constructor-parameter set, equipped with additional
structure when required. A Constructed Tensor Field Model (C-TFM) is the
augmented realization
\begin{align*}
  \hat{T}_{\mathrm C}
  =
  \bigl(\hat{T},\mathcal C,\Theta,\mathfrak E\bigr)
  \qquad
  \mathfrak E:
  \Theta\times\mathcal C
  \longrightarrow
  \mathfrak H^{(n)}
\end{align*}
For fixed \(\theta\in\Theta\), write
\begin{align*}
  E_\theta(c)
  :=
  \mathfrak E(\theta,c)
  =\mathbf H_c
\end{align*}
The requirement that \(E_\theta\) depend on \(c\) but not on the evaluation
query \((x,t)\) is part of the C-TFM structure.
Its conditional tangent section is the composition
\begin{align*}
  c
  \longmapsto
  \mathbf H_c=E_\theta(c)
  \longmapsto
  \widehat v_c
  =
  \mathfrak F_\phi^{(n)}[\mathbf H_c]
  \in\mathcal V
\end{align*}
The constructor may depend jointly on all parts of \(c\). If
\(\mathcal C=\prod_{i=1}^n\mathcal C_i\),
\(\Theta=\prod_{i=1}^n\Theta_i\), and
\(c=(c_1,\ldots,c_n)\), a Component-Separable Tensor Field Model (CS-TFM) is
the augmented realization
\begin{align*}
  \hat{T}_{\mathrm{CS}}
  =
  \bigl(\hat{T}_{\mathrm C},\{\mathfrak E_i\}_{i=1}^n\bigr)
\end{align*}
with specified component constructors
\begin{align*}
  \mathfrak E_i:
  \Theta_i\times\mathcal C_i
  \longrightarrow
  \mathcal H_i
  \qquad i=1,\ldots,n
\end{align*}
For fixed \(\theta=(\theta_1,\ldots,\theta_n)\), define the component Fields by
\begin{align*}
  H_{c_i}^{(i)}
  &:=
  E_{\theta_i}^{(i)}(c_i)
  :=
  \mathfrak E_i(\theta_i,c_i)
  \in\mathcal H_i,
  \qquad i=1,\ldots,n
\end{align*}
such that
\begin{align*}
  E_\theta(c)
  =
  \mathbf H_c
  =
  \left(H_{c_1}^{(1)},\ldots,H_{c_n}^{(n)}\right)
\end{align*}
In the conditional forward map, cross-component interactions can occur only
through \(\mathfrak F_\phi^{(n)}\), while the component families may have
different geometric types, resolutions, or latent representations. Forgetting
the componentwise factorization yields a C-TFM, and forgetting the remaining
construction data yields the underlying TFM realization:
\begin{align*}
  \text{CS-TFM}
  \longrightarrow
  \text{C-TFM}
  \longrightarrow
  \text{TFM}.
\end{align*}
These arrows schematically denote the corresponding forgetful maps between
structured realizations. 

\subsubsection{Tensor-Bundle Tensor Field Models}

For a fixed Base Manifold \(\Omega\), a \emph{Tensor Specification} is a
finite list
\begin{align*}
  \mathcal S
  =
  \{(C_a,r_a,s_a,V_a)\}_{a=1}^{m}
\end{align*}
Here \(m\geq1\), each \(C_a\to\Omega\) is a finite-rank carrier Bundle,
\(r_a,s_a\in\mathbb N_0\), and \(V_a\) is a finite-dimensional real
Vector Space. Write \(\underline V_a:=\Omega\times V_a\). The specification
determines the Tensor-structured Bundle
\begin{align*}
  \operatorname{Ten}(\mathcal S)
  &:=
  \bigoplus_{a=1}^{m}
  \left(
    \mathbb T^{r_a}_{s_a}(C_a)
    \otimes\underline V_a
  \right)
\end{align*}
The space \(V_a\) is a multiplicity space that allows several channels of the
same Tensor type.

\noindent A Tensor-Bundle Tensor Field Model (TB-TFM) is a TFM together with one Tensor
specification \(\mathcal S_i\) for each component and a Vector Bundle
isomorphism
\begin{align*}
  \hat{T}_{\mathrm{TB}}
  &:=
  \left(
    \hat{T},\{(\mathcal S_i,\tau_i)\}_{i=1}^n
  \right)
  &
  \tau_i:E_i
  &\overset{\sim}{\longrightarrow}
  \operatorname{Ten}(\mathcal S_i)
\end{align*}
where \(\tau_i\) covers \(\operatorname{id}_{\Omega_i}\). The admissible
family \(\mathcal H_i\) remains a specified subset of
\(\operatorname{Sec}(E_i)\).

\noindent Writing
\(\mathcal S_i=\{(C_{ia},r_{ia},s_{ia},V_{ia})\}_{a=1}^{m_i}\), this
definition separates the two roles clearly: \(C_{ia}\) supplies the Tensor
indices, while \(E_i\) is the component Bundle used by the TFM.
Ordinary Tensors on \(\Omega_i\) use \(C_{ia}=T\Omega_i\). Internal Tensors
may instead use \(C_{ia}=\Omega_i\times W_{ia}\), or any other prescribed
carrier Bundle over \(\Omega_i\). In each summand the Tensor slots transform
under changes of frame of \(C_{ia}\), while the multiplicity coordinates in
\(V_{ia}\) are unchanged.

\noindent A TB-TFM is a TFM with additional
geometric typing data, not necessarily a smaller class of underlying models.
The typing alone imposes no symmetry condition on the constructors or Field
Operator.

\subsubsection{Hierarchy Overview}

The hierarchy consists of one construction branch and one compatible
Tensor Bundle refinement. Tensor Bundle structure adds Tensor specifications
and Bundle identifications and can be combined with a TFM, C-TFM, or CS-TFM.
Combined names concatenate only nonredundant prefixes: because every CS-TFM is
already a C-TFM, for example, a component-separable Tensor Bundle realization
is called a CS-TB-TFM.

\begin{equation*}
  \text{TFM}
  \;\supset\;
  \begin{cases}
    \text{C-TFM}\supset\text{CS-TFM},\\
    \text{TB-TFM}.
  \end{cases}
\end{equation*}
Here \(\supset\) denotes the hierarchy relation ``is more general than'':
forgetting the additional data and compatibility conditions of a realization
on the right yields a realization of the class on the left. It is not meant as
literal set inclusion, nor must the relation be proper. The Tensor Bundle
refinement can be combined with either construction class when the required
specifications and Bundle identifications are supplied. Tangent-output
compatibility and membership in \(\mathcal V\) belong to the root definition,
and exact reuse follows from a condition-only constructor, so neither forms a
separate refinement. ODE-admissibility and other analytic restrictions are
determined by the chosen output family relative to the declared solution
concept.

\subsection{Amortized Point Evaluation}
\label{sec:tfm-amortization}

For a C-TFM, the constructor depends on \(c\) but not on the query
\((x,t)\). Hence a batch
\(\mathcal B_c=\{q_j=(x_j,t_j)\}_{j=1}^{N_q}\) sharing the same condition also shares the
single Field collection \(\mathbf H_c\):
\begin{align*}
  \widehat v_c(x_j,t_j)
  =
  F_\phi(\mathbf H_c,x_j,t_j)\in T_{x_j}\mathcal M
  \qquad j=1,\ldots,N_q
\end{align*}
This is a computational consequence of the factorization, not an additional
geometric refinement. It supports batched or parallel evaluation. The
following idealized additive cost model assumes query-independent costs. Let
\(C_{\mathrm{build}}\geq0\) be the cost of constructing \(\mathbf H_c\),
\(C_F>0\) the cost of one point evaluation, and \(C_G>0\) the cost of one
query to an unfactorized predictor. Then
\begin{align*}
  C_{\mathrm{TFM}}(N_q)
  =
  C_{\mathrm{build}}+N_q C_F
  \qquad
  C_{\mathrm{base}}(N_q)
  =
  N_q C_G
\end{align*}
For \(N_q\in\mathbb N\), reuse is beneficial precisely when
\begin{align*}
  C_{\mathrm{build}}+N_q C_F<N_q C_G
  \qquad\text{equivalently}\qquad
  \frac{C_{\mathrm{build}}}{N_q}+C_F<C_G
\end{align*}
If \(C_G\leq C_F\), this inequality cannot hold. If \(C_G>C_F\), it is
equivalent to the crossover condition
\begin{align*}
  N_q>\frac{C_{\mathrm{build}}}{C_G-C_F}
\end{align*}
The corresponding idealized speedup is
\begin{align*}
  S_{N_q}
  =
  \frac{N_q C_G}{C_{\mathrm{build}}+N_q C_F}
  \longrightarrow
  \frac{C_G}{C_F}
\end{align*}
as \(N_q\) grows. Thus, construction is amortized across the shared queries,
while the large-fan-out speedup is bounded by the relative per-query costs.

\subsection{Training Objectives}
Here, \(\mathbb E\) denotes expectation under task-specific probability laws on
the relevant measurable spaces. Every displayed random quantity and loss
integrand is assumed measurable and integrable whenever the corresponding loss
is used.
To define the training objectives for constructed Tensor Field Models, equip
\(\mathcal M\) with a Riemannian metric \(g\) and define
\begin{align*}
  \ell_x(u,v)=g_x(u-v,u-v),
  \qquad u,v\in T_x\mathcal M
\end{align*}
For each condition \(c\), construct \(\mathbf H_c=E_\theta(c)\) once and share
it across the \(N_q\) state--time queries. Given target Vectors
\(v_j^\star\in T_{x_j}\mathcal M\), direct training minimizes
\begin{align*}
  \mathcal L_{\mathrm{task}}^{(n)}
  =
  \mathbb E\!\left[
    \frac{1}{N_q}\sum_{j=1}^{N_q}
    \ell_{x_j}\!\left(
      F_\phi(\mathbf H_c,x_j,t_j),v_j^\star
    \right)
  \right]
\end{align*}
This preserves the model factorization: \(\mathbf H_c\) depends only on the
shared condition, whereas \(F_\phi\) additionally receives the query
\((x_j,t_j)\).

\noindent For post-training conversion, let \(G_{\mathrm{ref}}\) denote a
reference Vector Field satisfying
\begin{align*}
  G_{\mathrm{ref}}(x,t,c)\in T_x\mathcal M
\end{align*}
For \(c\sim\nu\) and \((x,t)\sim\mu_c\), pointwise distillation minimizes
\begin{align*}
  \mathcal L_{\mathrm{field}}^{(n)}
  =
  \mathbb E\!\left[
    \ell_x\!\left(
      F_\phi(\mathbf H_c,x,t),G_{\mathrm{ref}}(x,t,c)
    \right)
  \right]
\end{align*}
Because both Vector Fields lie in \(T_x\mathcal M\), no transport between
tangent spaces is required. Optional rollout distillation generates reference
and TFM trajectories using the same condition, initial state, time grid, and
update noise, together with a Manifold-preserving numerical update of the form
\begin{align*}
  \Psi_\ell:
  \{(x,u):x\in\mathcal M,\ u\in T_x\mathcal M\}
  \times\Xi\times\mathcal I^2
  \longrightarrow\mathcal M
\end{align*}
Using rollouts presupposes that the selected outputs are admissible for the
chosen update rule. A continuous-time ODE interpretation additionally requires
the TFM and reference outputs to be ODE-admissible under the declared solution
concept.
Let \(t_0<\cdots<t_K\) be a time grid in \(\mathcal I\), fix
\(x_0\in\mathcal M\), set
\(x_{t_0}^{\mathrm{TFM}}=x_{t_0}^{\mathrm{ref}}=x_0\), and use the same noise
variables \(\xi_0,\ldots,\xi_{K-1}\in\Xi\) in both rollouts. For
\(\ell=0,\ldots,K-1\), define
\begin{align*}
  x_{t_{\ell+1}}^{\mathrm{TFM}}
  &=
  \Psi_\ell\!\left(
    \left(
      x_{t_\ell}^{\mathrm{TFM}},
      F_\phi(
        \mathbf H_c,x_{t_\ell}^{\mathrm{TFM}},t_\ell
      )
    \right),
    \xi_\ell,
    (t_\ell,t_{\ell+1})
  \right)\\
  x_{t_{\ell+1}}^{\mathrm{ref}}
  &=
  \Psi_\ell\!\left(
    \left(
      x_{t_\ell}^{\mathrm{ref}},
      G_{\mathrm{ref}}(
        x_{t_\ell}^{\mathrm{ref}},t_\ell,c
      )
    \right),
    \xi_\ell,
    (t_\ell,t_{\ell+1})
  \right)
\end{align*}
Using the geodesic distance \(d_g\), the rollout objective is
\begin{align*}
  \mathcal L_{\mathrm{roll}}^{(n)}
  =
  \mathbb E\!\left[
    \sum_{\ell=1}^{K}\alpha_\ell
    d_g\!\left(x_{t_\ell}^{\mathrm{TFM}},x_{t_\ell}^{\mathrm{ref}}\right)^2
  \right]
  \qquad \alpha_\ell\geq0
\end{align*}

\noindent A Field regularizer
\begin{align*}
  \mathcal L_H^{(n)}
  =
  \mathbb E[\mathcal R(\mathbf H_c)]
\end{align*}
may be used to control the norm, smoothness, or redundancy of the component
Fields.
The complete objective is
\begin{align*}
  \mathcal L^{(n)}
  =
  \lambda_{\mathrm{task}}\mathcal L_{\mathrm{task}}^{(n)}
  +\lambda_{\mathrm{field}}\mathcal L_{\mathrm{field}}^{(n)}
  +\lambda_{\mathrm{roll}}\mathcal L_{\mathrm{roll}}^{(n)}
  +\lambda_H\mathcal L_H^{(n)}
  \qquad
  \lambda_{\mathrm{task}},\lambda_{\mathrm{field}},
  \lambda_{\mathrm{roll}},\lambda_H\geq0
\end{align*}
Terms that are not used are assigned zero weight.

\newpage
\section{Experiments}
\label{sec:experiments}

Three consequences of the preceding formulation are tested. First, a restricted
component factorization can improve generalization when the conditional family
has matching structure. Second, a typed Field Operator can impose a known
symmetry exactly. Third, a condition-only construction can be reused without
changing the output. Every learned model below is trained with the same Flow
Matching (FM) or Riemannian Flow Matching (RFM) objective as its control
\cite{lipman2023flowmatching,chen2024riemannianfm}. TFM and direct FM
therefore distinguish Vector-Field parameterizations under a common objective.

\noindent Learned-model results use five paired seeds. Reported values are the seed mean
and a two-sided \(95\%\) Student-\(t\) confidence interval with four degrees of
freedom. Paired percentage reductions are computed within a seed before
aggregation. Timing results instead report medians and interquartile ranges
over randomized timing blocks.

\subsection{Compositional Conditional Flow Matching in \texorpdfstring{
\(\mathbb R^4\)}{R4}}

\paragraph{Task.}
Let \(c=(i,j)\), with \(i,j\in\{1,\ldots,8\}\), and
\begin{align*}
  p_0&=\mathcal N(0,I_4)
  &
  p_1(x\mid i,j)&=\mathcal N(\mu_{ij},\sigma^2 I_4)
  &
  \sigma&=0.35
\end{align*}
In the compositional task, the means are generated by a bilinear contraction
with latent dimension three in both condition modes:
\begin{align*}
  \mu_{ij,d}
  =
  \sum_{r=1}^{3}\sum_{s=1}^{3}
  u_{ir}A_{drs}w_{js}
  \qquad d=1,\ldots,4
\end{align*}
Consequently, the resulting \(8\times8\times4\) Tensor has row and column
multilinear ranks at most three. This construction does not assert CP rank
three.
The means are then scaled using training combinations only to have coordinate RMS
\(1.5\). A fixed perfect matching of the \(8\times8\) condition table is held
out. The remaining \(56\) pairs form a connected bipartite graph, and every
row and column value occurs in seven training pairs. Evaluation therefore
tests eight unseen combinations of observed components.

\noindent Both models use the independent straight CFM coupling
\begin{align*}
  x_0&\sim\mathcal N(0,I_4)
  &x_1&=\mu_{ij}+\sigma\epsilon
  &x_t&=(1-t)x_0+t x_1
  &u_t&=x_1-x_0
\end{align*}
where \(t\sim\operatorname{Unif}[0,1]\) and
\(\epsilon\sim\mathcal N(0,I_4)\). In addition to generated samples, evaluation
uses the analytic marginal regression Field
\begin{align*}
  v_{\mathrm{marg}}(x,t,i,j)
  =
  \mu_{ij}
  +
  \frac{\sigma^2t-(1-t)}{(1-t)^2+\sigma^2t^2}
  (x-t\mu_{ij})
\end{align*}

\paragraph{Models and protocol.}
The CS-TFM constructors return separate learned representations
\(h_i,g_j\in\mathbb R^3\). Only the Field Operator contracts them:
\begin{align*}
  z_k(i,j)
  &=
  \sum_{a,b=1}^{3}C_{kab}h_{i,a}g_{j,b}
  \qquad k=1,\ldots,8
  \\
  q(x,t)&=\operatorname{MLP}_{32}(x,t)
  &
  \widehat v_d
  &=
  [Bq]_d+\sum_{h,k}T_{dhk}q_hz_k
\end{align*}
The query network has two width-\(32\) SiLU layers and \(1{,}248\) parameters.
The complete CS-TFM has \(2{,}524\) parameters. The direct conditional FM
control is a two-hidden-layer, width-\(38\) SiLU MLP applied to
\([x,t,\operatorname{onehot}(i),\operatorname{onehot}(j)]\). It has
\(2{,}474\) parameters, \(2.0\%\) fewer. Its first affine layer already
contains separate row and column embeddings implicitly, while its subsequent
layers are not restricted to a Tensor contraction.

\noindent Within a seed, the models receive identical minibatches and evaluation noise.
Training uses \(4{,}000\) AdamW updates, batch size \(512\), learning rate
\(2\times10^{-3}\), weight decay \(10^{-6}\), gradient clipping at \(5\), and
cosine learning-rate decay to \(2\times10^{-4}\). Evaluation uses \(1{,}024\)
velocity queries and \(512\) generated samples per condition. Generation uses
\(64\) Heun steps, and sliced \(W_2\) uses \(128\) random directions and exact
Gaussian target quantiles.

\paragraph{Results and falsification controls.}
On the held-out bilinear combinations, the CS-TFM reduces sliced \(W_2\) by
\(82.9\%\) (paired \(95\%\) CI \(68.6\%\)--\(97.3\%\)). Every seed favors the
CS-TFM. Its train and held-out analytic velocity NMSEs are respectively
\(0.0183\) and \(0.0191\), whereas the direct model changes from \(0.0234\) on
training combinations to \(0.3717\) on held-out combinations. This is a
compositional-generalization result, not merely a better fit to observed
pairs.

\noindent For a falsification control, the bilinear Tensor is replaced by an IID Gaussian
\(8\times8\times4\) mean table, and the complete protocol is repeated. An unseen
entry then contains no recoverable row--column information. The CS-TFM no
longer improves generation: its sliced \(W_2\) is \(2.022\), versus \(1.944\)
for direct FM, a paired reduction of \(-4.1\%\)
\([-15.6\%,7.5\%]\). It also fits the observed IID table much less accurately
(training velocity NMSE \(0.325\), versus \(0.0264\)), as expected from its
latent-dimension restriction.

\noindent The main model deliberately matches the simulator's latent dimension so that
the controlled test isolates the proposed inductive bias. A latent-dimension
sensitivity check makes this dependence explicit. Mild over-specification with
model latent dimension four retains a
\(79.1\%\) sliced-\(W_2\) reduction
\([63.3\%,95.0\%]\), while latent dimension two gives no reliable benefit
(\(-59.0\%\), \([-143.0\%,25.0\%]\)).

\subsection{Exactly Equivariant RFM on \texorpdfstring{
\(\mathbb S^2\)}{S2}}

\paragraph{Task.}
Conditions are \(c=(a,b)\in\mathbb S^2\times\mathbb S^2\). Training uses the
canonical orientation
\begin{align*}
  a&=e_3
  &
  b&=(\sin\varphi,0,\cos\varphi)
  &
  \varphi&\sim\operatorname{Unif}[35^\circ,85^\circ]
\end{align*}
and the endpoint law
\begin{align*}
  p_1(x\mid a,b)
  =
  \operatorname{vMF}\!\left(
    \frac{a+0.75b+0.35(a\times b)}
         {\|a+0.75b+0.35(a\times b)\|},
    18
  \right)
\end{align*}
For \(x_0\sim\operatorname{Unif}(\mathbb S^2)\), \(x_1\sim p_1\), and
\(\vartheta=\arccos\langle x_0,x_1\rangle\), one has
\(\vartheta\in(0,\pi)\) almost surely under these continuous laws. On this
event, let \(d=(x_1-\cos\vartheta\,x_0)/\sin\vartheta\). The shortest-geodesic
CFM path and its tangent target are
\begin{align*}
  x_t
  &=\cos(t\vartheta)x_0+\sin(t\vartheta)d
  &
  u_t
  &=\vartheta[-\sin(t\vartheta)x_0+\cos(t\vartheta)d]
\end{align*}
The implementation supplies a stable tangent direction in the probability-zero
coincident and antipodal cases.

\paragraph{Models and protocol.}
Write \(P_x=I-xx^\top\). The CS-TB-TFM constructs typed component Fields from
\(a\) and \(b\), and its shared Operator multiplies the five tangent basis
Fields
\begin{align*}
  \mathcal B(x,a,b)
  =
  \bigl(P_xa,P_xb,P_x(a\times b),x\times a,x\times b\bigr)
\end{align*}
by coefficients from a width-\(96\), three-hidden-layer SiLU MLP. That MLP
receives only
\begin{align*}
  \bigl(t,\langle x,a\rangle,\langle x,b\rangle,
  \langle a,b\rangle,\langle x,a\times b\rangle\bigr)
\end{align*}
All inputs to the coefficient map are invariant, while the basis Fields are
covariant. Hence, for every parameter value and \(R\in\mathrm{SO}(3)\),
\begin{align*}
  \widehat v(Rx,t,Ra,Rb)=R\widehat v(x,t,a,b)
  \qquad
  \widehat v(x,t,a,b)\in T_x\mathbb S^2
\end{align*}
The model has \(19{,}685\) parameters.

\noindent The control is a width-\(95\), three-hidden-layer coordinate MLP on
\([x,t,a,b]\), followed by the same projection \(P_x\). It has \(19{,}573\)
parameters (\(0.57\%\) fewer). Thus tangency is not a confound. The CS-TB-TFM
and one coordinate-MLP control are trained on canonical examples without
augmentation. A second coordinate-MLP control, initialized identically to the
first, receives a fresh Haar rotation applied jointly to \((x,a,b,u_t)\) on
every training example. All three models share the underlying minibatch stream.
Training uses \(3{,}000\) AdamW
updates, batch size \(512\), learning rate \(2\times10^{-3}\), weight decay
\(10^{-6}\), and gradient clipping at \(10\). Evaluation uses \(8{,}192\)
paired canonical examples and jointly Haar-rotated copies. The rotated raw
inputs lie outside the sampled canonical training support of both the CS-TB-TFM
and the canonical-only coordinate MLP, and they are in distribution for the
augmented control. For the CS-TB-TFM, however, rotation leaves the invariant
coefficient inputs unchanged and transforms the basis Fields covariantly, so
its behavior on the rotated inputs is fixed exactly by equivariance. Generation
uses \(2{,}048\) paired initial states and \(48\) embedded Heun steps with
normalization retraction.

\paragraph{Results.}
Canonical velocity NMSE is essentially equal for the CS-TB-TFM and
canonical-only coordinate MLP: \(0.2983\) and \(0.2988\), respectively. After
Haar rotation, the TFM mean remains \(0.2983\), with canonical and rotated NMSE
equal within every seed up to numerical precision, while the raw coordinate
MLP is \(5.9\)--\(50.2\) times worse seedwise. The geometric-mean
error ratio is \(12.9\) (log-ratio \(95\%\) CI \(4.0\)--\(41.4\)). The TFM's
normalized equivariance defect is \(3.0\times10^{-14}\).

\noindent Haar augmentation largely closes the gap. Nevertheless, at the same update
budget the CS-TB-TFM has \(5.92\%\) lower Haar-rotated velocity NMSE than the
augmented control (paired \(95\%\) CI \(4.94\%\)--\(6.89\%\)). Their generated
distributions are comparable: axial \(W_1\) is \(0.0060\) for the TFM and
\(0.0083\) for the augmented control, whereas it is \(0.573\) for the rotated
canonical-only MLP. The result is therefore interpreted as exact rotational
generalization and a fixed-budget inductive-bias benefit, not as proof that
augmentation or another equivariant architecture cannot match it.

\begin{table}[h!]
\centering
\small
\caption{Primary held-out results. Model entries are mean
\([95\%\text{ Student-}t\text{ CI}]\) across five seeds. The last column is
the mean seedwise percentage reduction of the TFM relative to its named
control. The Euclidean control is direct FM. The spherical control is the
Haar-augmented projected MLP. Lower is better.}
\label{tab:tfm-primary-results}
\resizebox{0.98\textwidth}{!}{%
\begin{tabular}{lllll}
\hline
Setting & Metric & Structured TFM & Control & Paired reduction \\
\hline
Bilinear \(\mathbb R^4\)
& sliced \(W_2\)
& \(0.0549\,[0.0480,0.0617]\)
& \(0.447\,[0.116,0.779]\)
& \(82.9\%\,[68.6\%,97.3\%]\) \\
IID \(\mathbb R^4\) control
& sliced \(W_2\)
& \(2.022\,[1.599,2.446]\)
& \(1.944\,[1.610,2.278]\)
& \(-4.1\%\,[-15.6\%,7.5\%]\) \\
Haar-rotated \(\mathbb S^2\)
& velocity NMSE
& \(0.2983\,[0.2946,0.3020]\)
& \(0.3171\,[0.3100,0.3241]\)
& \(5.92\%\,[4.94\%,6.89\%]\) \\
\hline
\end{tabular}%
}
\end{table}

\begin{figure}[h!]
  \centering
  \includegraphics[width=\textwidth]{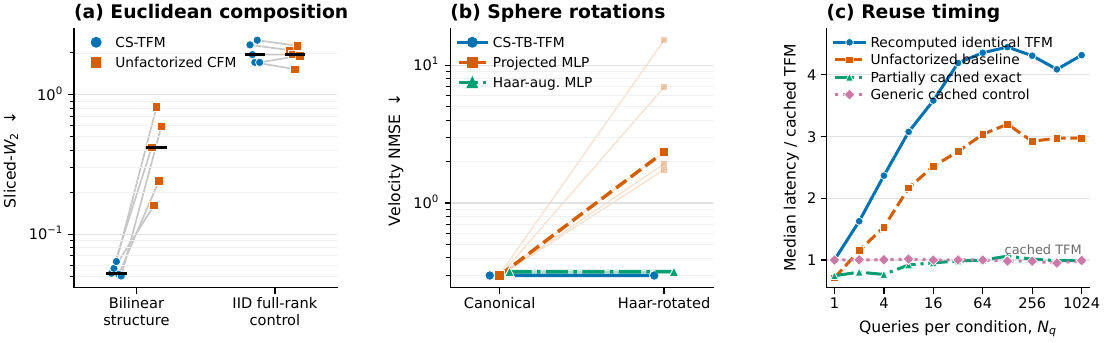}
  \caption{Controlled experimental summary. (a) Seedwise held-out sliced
  \(W_2\). Gray segments pair the same seed and black bars mark medians. The
  CS-TFM advantage on the bilinear family disappears for the IID control.
  (b) Canonical and jointly Haar-rotated velocity NMSE. All models are tangent.
  Pale lines are individual seeds and heavy lines are medians. The rotated raw
  inputs lie outside the sampled canonical training support of both the
  CS-TB-TFM and the unaugmented coordinate MLP. In the CS-TB-TFM, rotation
  leaves the coefficient inputs invariant and transforms the basis Fields
  covariantly. The augmented MLP sees such rotations during training. (c) Median
  serial CPU latency divided by cached-TFM latency. Rebuilding the identical
  TFM or recomputing the direct predictor is costly, but a generic cached graph
  and an exactly partially cached MLP match the TFM, showing that reuse is not
  unique to the terminology.}
  \label{fig:tfm-experiments}
\end{figure}

\subsection{Condition-Reuse Audit}

The additive law in \autoref{sec:tfm-amortization} is tested separately with
small deterministic scalar Python MLPs on one Apple-arm64 CPU. The TFM has
\(1{,}164\) parameters, \(768\) construction multiply--accumulates (MACs), and
\(300\) Operator MACs per query. A parameter-matched direct MLP has \(1{,}180\)
parameters and \(1{,}148\) MACs per complete query. For
\(N_q\in\{1,2,4,\ldots,1024\}\), the timing protocol uses \(31\)
randomized-order repetitions, five warmups, and disabled garbage collection
inside timed blocks.

\noindent At \(N_q=1024\), building the TFM once is \(4.32\times\) faster than rebuilding
the identical TFM at every query and \(2.97\times\) faster than recomputing the
direct MLP. Cached and recomputed outputs agree exactly. These comparisons
alone would overstate the consequence, however. A generic conditioner with the
identical TFM graph has the same latency, and exactly caching the direct MLP's
first-layer contribution \(b+W_c c\) also preserves its output bitwise. Its
cost is \(896+252N_q\) MACs, versus \(768+300N_q\) for the TFM, so the partially
cached ordinary predictor is asymptotically cheaper and is \(0.989\) times the
TFM latency at \(N_q=1024\) in this run. Thus the cost law and exact reuse are
verified, but speed is architecture-, implementation-, and workload-dependent
and is not unique to TFMs. The scalar CPU audit is not an end-to-end sampler
benchmark.

\paragraph{Independent-path batching.}
An additional systems throughput audit evaluates independent conditional
trajectories using the preceding equivariant TFM (19,685 parameters) and the
direct projected MLP (19,573 parameters) on an Apple M3 CPU (eight cores,
8\,GB RAM) with PyTorch 2.6, four intra-op threads, eager FP32 execution, and
fixed, deterministically initialized (untrained) weights. For each
\(B\in\{1,8,32,128,512\}\), both models receive the same seeded initial
states under one fixed condition and integrate them either with an outer
Python loop over singleton paths or as one tensor batch. Both use 48 Heun
steps (96 function evaluations). The integration steps themselves remain
sequential. The condition cache is prepared before the timed integration for
both the equivariant TFM and an exactly refactored direct MLP, so this audit
measures path batching rather than cache construction. Reported values are
medians and inclusive interquartile ranges over 15 randomized-order blocks
after three warmups.

\begin{table}[h!]
\centering
\small
\caption{CPU latency for independent-path sampling. Times are median
\([\mathrm{IQR}]\) in milliseconds. Speedup is the ratio of serial and
batched medians within each model.}
\label{tab:parallel-sampling}
\resizebox{\textwidth}{!}{%
\begin{tabular}{crrrrrr}
\hline
\(B\)
& TFM serial ms & TFM batched ms & TFM speedup
& Direct serial ms & Direct batched ms & Direct speedup \\
\hline
1
& \(6.57\,[5.63,8.23]\) & \(5.83\,[5.61,8.01]\) & \(1.13\times\)
& \(3.99\,[3.91,4.15]\) & \(4.04\,[3.87,4.37]\) & \(0.99\times\) \\
8
& \(52.06\,[47.37,54.71]\) & \(7.88\,[7.16,8.93]\) & \(6.61\times\)
& \(32.20\,[31.58,33.10]\) & \(5.04\,[4.99,5.27]\) & \(6.39\times\) \\
32
& \(242.72\,[215.22,311.89]\) & \(9.60\,[8.69,13.43]\) & \(25.28\times\)
& \(129.13\,[128.45,129.85]\) & \(6.34\,[6.25,6.56]\) & \(20.38\times\) \\
128
& \(1082.72\,[840.44,1161.49]\) & \(16.32\,[14.21,23.34]\) & \(66.33\times\)
& \(528.81\,[520.16,535.19]\) & \(10.41\,[10.29,10.57]\) & \(50.82\times\) \\
512
& \(2923.10\,[2880.50,3010.21]\) & \(25.92\,[25.50,27.38]\) & \(112.76\times\)
& \(2103.91\,[2069.18,2165.00]\) & \(20.75\,[19.72,23.60]\) & \(101.41\times\) \\
\hline
\end{tabular}%
}
\end{table}

\noindent At \(B=512\), vectorization is \(112.76\times\) faster for the TFM and
\(101.41\times\) faster for the direct MLP. The
direct MLP also has lower absolute batched latency at every reported
\(B\). The serial and batched endpoints agree to a maximum absolute
discrepancy of \(1.2\times10^{-7}\). These measurements characterize dense
execution throughput only and do not compare sample quality.

\section{Related Work}
\label{sec:related-work}
A useful intuition is to regard the components of \(c\) as \(n\)
condition-dependent sources. Component \(c_i\) determines a typed Field
\(H_{c_i}^{(i)}\), whose information is available wherever that Field is
evaluated. At a query \((x,t)\), the learned Operator combines the complete
collection \(\mathbf H_c\) and returns a velocity in \(T_x\mathcal M\).
This retains the ``propagated information'' picture, but makes two distinctions
precise: the components \(c_i\) need not themselves be points of
\(\mathcal M\), and in the architectures studied here no single
\(H_{c_i}^{(i)}\) is the generated tangent section. The root definition does
not forbid the latter degenerate case.
The closest connections are therefore to Operator learning, conditional
function representations, geometric deep learning, and continuous-time
generative models.

\paragraph{Operator learning and conditional function representations.}
Neural Operators learn maps between function spaces rather than only maps
between finite-dimensional vectors~\cite{kovachki2023neuraloperator}.
DeepONet separates an encoding of an input function from the coordinates at
which the output function is queried~\cite{lu2021deeponet}, while Fourier
Neural Operators parameterize global integral Operators efficiently in the
Fourier domain~\cite{li2021fno}. TFMs share this function-to-function
viewpoint and, in their point-evaluation form, the separation between a
condition-dependent representation and an output coordinate. Their emphasis
is different: the input is a product of potentially heterogeneous section
families, the output belongs to the prescribed family
\(\mathcal V\subseteq
\operatorname{Sec}(\operatorname{pr}_{\mathcal M}^{*}T\mathcal M)\), and the
component Fields are constructed once for repeated queries. The present
formulation does not by itself imply the discretization invariance or
resolution transfer often sought in Neural Operator architectures. Those
properties depend on the chosen constructors and Field Operator.

\paragraph{Continuous-time generative Vector Fields.}
Neural ordinary differential equations and continuous normalizing flows
established learnable dynamics defined by neural Vector
Fields~\cite{chen2018neuralode}. Flow Matching trains such Fields by regression
against conditional probability paths~\cite{lipman2023flowmatching}. Diffusion
and score-based models provide closely related iterative dynamics in discrete,
ODE, or SDE form~\cite{ho2020ddpm,song2021ddim,song2021sde}. On Manifolds,
Riemannian score models and Riemannian Flow Matching ensure that the learned
dynamics respect the state geometry~\cite{debortoli2022riemanniansgm,
chen2024riemannianfm}. TFMs are complementary to these training paradigms:
they specify how a structured condition parameterizes the Vector Field, not
which probability path or stochastic process supplies the regression target.
Consequently, the same factorization can be trained directly, distilled from
an existing Field, or combined with rollout supervision.

\paragraph{Tensor-valued and equivariant networks.}
Tensor Field Networks process scalar, vector, and higher-order features with
rotation- and translation-equivariant operations on point
clouds~\cite{thomas2018tfn}. Despite the similar name, they are not the same
construction as Tensor Field Models. Gauge-equivariant convolutional networks
extend typed feature Fields to Manifolds while remaining independent of local
frame choices~\cite{CohenGaugeCNN}, and geometric deep learning more broadly
organizes architectures around symmetries and domain
structure~\cite{BronsteinGeometricDL}. TFMs adopt the compatible principle that
intermediate and output quantities should have explicit geometric types, but
focus on factorizing a condition into a Field collection and then a Vector
Field.
Equivariance is guaranteed only when the component constructors, contractions,
and Field Operator are chosen equivariantly, as in the \(\mathbb S^2\)
experiment. It is not automatic for every TFM parameterization.

\paragraph{Reduced representations and inference acceleration.}
When explicit capacity restrictions are imposed, the use of the intermediate
collection is related in spirit
to reduced-order modeling for parametric dynamical
systems~\cite{benner2015modelreduction}. It should not, however, be identified
with a classical matrix or multilinear low-rank
factorization~\cite{eckart1936,desilva2008tensor}: the component Fields may
live in different Bundles, and their learned composition is generally
nonlinear.

\noindent Several complementary approaches reduce the cost of iterative generation.
Fast ODE solvers reduce the number of Field
evaluations~\cite{lu2022dpmsolver}. Progressive distillation and consistency
models learn few-step or one-step generators~\cite{salimans2022progressive,
song2023consistency}. Parallel sampling reduces sequential
dependence~\cite{shih2023paradigms}. Feature-caching methods instead exploit
similarity between neighboring denoising
evaluations~\cite{ma2024deepcache,ma2024learningtocache,liu2025teacache,
gao2026meancache}. TFM reuse is orthogonal to these methods. It caches a
condition-only Field collection exactly across sampler steps and independent
paths, rather than approximately reusing state-dependent activations across
time. It therefore does not reduce the number of causal solver steps, and its
benefit emerges only when the construction cost is amortized over enough
shared-condition queries. This distinction also allows TFMs to be combined
with faster solvers, distillation, temporal caching, or path-level batching.
As the cache audit in \autoref{sec:experiments} demonstrates, the same exact
reuse is available to an ordinary predictor whenever its computation graph
exposes a condition-only subgraph. The TFM formulation makes this separation
explicit but does not monopolize it.

\nocite{Relational_Transformer}
\printbibliography

\newpage
\appendix

\end{document}